\documentclass[bibtex, nonblind]{apsr_submission}

\usepackage{latexsym}

\usepackage{graphicx} %
\usepackage{amsmath}
\usepackage{multirow}
\usepackage{array}
\usepackage{enumitem}
\usepackage{color, soul}
\usepackage{algorithm}
\usepackage{algpseudocode}
\usepackage{tcolorbox}

\usepackage{subcaption}
\usepackage{bbm}
\usepackage{wrapfig}

\title{Towards Coding Social Science Datasets with Language Models}
\author{Christopher Michael Rytting}
{Computer Science Department, Brigham Young University}
{Correspondence should be directed to christophermichaelrytting@gmail.com}

\author{Taylor Sorensen}
{Computer Science Department, Brigham Young University}

\author{Lisa Argyle}
{Political Science Department, Brigham Young University}

\author{Ethan Busby}
{Political Science Department, Brigham Young University}

\author{Nancy Fulda}
{Computer Science Department, Brigham Young University}

\author{Josh Gubler}
{Political Science Department, Brigham Young University}

\author{David Wingate}
{Computer Science Department, Brigham Young University}

\runningtitle{Towards Coding Social Science Datasets with Language Models}
\runningauthor{Rytting et al.} 

\begin{document}
\begin{frontmatter}
\begin{abstract}
Researchers often rely on humans to code (label, annotate, etc.) large sets of texts. This kind of human coding forms an important part of social science research, yet the coding process is both resource intensive and highly variable from application to application. In some cases, efforts to automate this process have achieved human-level accuracies, but to achieve this, these attempts frequently rely on thousands of hand-labeled training examples, which makes them inapplicable to small-scale research studies and costly for large ones. Recent advances in a specific kind of artificial intelligence tool - language models (LMs) - provide a solution to this problem. Work in computer science makes it clear that LMs are able to classify text, without the cost (in financial terms and human effort) of alternative methods. To demonstrate the possibilities of LMs in this area of political science, we use GPT-3, one of the most advanced LMs, as a synthetic coder and compare it to human coders.  We find that GPT-3 can match the performance of typical human coders and offers benefits over other machine learning methods of coding text. We find this across a variety of domains using very different coding procedures. This provides exciting evidence that language models can serve as a critical advance in the coding of open-ended texts in a variety of applications.
\end{abstract}
\end{frontmatter}

\section{Introduction}

The analysis of textual data--from sources like open-ended survey responses, social media posts, and legislative transcripts--has become increasingly important across many disciplines. Traditionally, researchers quantitatively analyzing these text have trained research assistants (mostly undergraduate students) to \textit{code} the material by assigning numbers and/or categories to text segments. However, such human coding is slow and expensive. Given variability in experience and perception among coders, researchers hire multiple people to evaluate the same texts when possible and then calculate intercoder agreement as a measure of confidence in the coding process. At times, even this repeated coding is not feasible, and researchers rely on a single human coder.

While this approach works for small amounts of text, it becomes impractical as a means to to analyze the texts available in an increasingly information-rich world. As a result, many scholars seek automated alternatives. Dictionary-based methods \citep{roberts2020linking, young2012affective} work in cases where clearly defined sets of words indicate the presence of particular content but struggle with nuance and generalization \citep{boydstun2021, grimmer2013text} 
One solution to this problem uses supervised machine learning (SML) models to code text in the place of humans, such as naive bayes, random forests, and SVMs \citep{grimmer2013text, boydstun2021}. Unfortunately, all of these require large datasets for training, which typically must be hand-generated by human coders, failing to eliminate the time and expense of using human coders \citep{collingwood2012tradeoffs}. SML methods also require large datasets with a sufficient sample size to train, test, and validate a SML procedure. Unsupervised methods exist - such as structural topic modeling \citep{roberts2014} - but these still require significant amounts of data and extensive modeling and validation steps. Most importantly, they do not allow researchers to intentionally code specific themes and topics.

We propose that state-of-the-art artificial intelligence tools, known as language models (LMs), provide a powerful alternative to current techniques for coding texts in the social sciences, as has been done in labeling in other domains and methodologies including stance detection, psychology, and synthetic dataset generation \citep{burnham2023stance, rathje2023gpt, gilardi2023chatgpt, halterman2023synthetically}. We describe these tools and the application of one - GPT-3 \citep{brown2020language} - to various coding tasks in political science. We show that GPT-3 performs coding tasks at or exceeding the level of human coders, even when it is given three or fewer labeled examples. We also find that GPT-3 performs comparably to SML procedures, with a fraction of the time and cost of those approaches.

\section{Language models}



In the most basic sense, LMs are a conditional probability distribution $p(x_n|x_1,\cdots,x_{n-1})$ over tokens or words. LMs generate novel sequences of text by repeatedly sampling from this distribution. Crucially, LMs can be given initial inputs that reduce the probability of some output statements and increase the probability of others. Given the initial input of ``Will you please'', a LM might assign high probability to ``go'' as the next term, and low probability to ``fruit''. Changing the context to ``Will you eat'' switches those probabilities. 

The use of LMs in social science has recently seen much progress and promise \citep{KrepsMcCainBrundage2020, Arglyeetal, argyle2023ai, bail2023can, ziems2023can}. LMs can serve as useful tools in coding texts for at least two reasons. First, LMs are created and trained on massive amounts of human created statements. This means the models come already set up with an extensive understanding of human texts. Second (and relatedly), LMs have few-shot capabilities or the capacity to learn complicated tasks with only a handful of examples. This can almost entirely eliminate the need for hand-coded training data, providing advantages even over SML methods.

For our application here, we use GPT-3, one of the largest existing LMs. This language model was released by OpenAI in 2020, has 175 billion parameters, and was trained on more than 45 terabytes of text. In automated content analysis, others have considered different, custom-modified LMs such as BERT \citep{devlin-etal-2019-bert}, BART \citep{lewis-etal-2020-bart}, RoBERTa \citep{liu2019roberta}, XLNet \citep{yang2019xlnet}, and ELMo \citep{Peters2018}. However, these all require extensive fine-tuning and a similar number of labeled examples as SML methods. As such, we explore GPT-3 as a coding tool with only few-shot learning methods (and no fine tuning) to determine if it provides a more efficient automated coding tool that is more accessible to most social science researchers. 


\section{Methodology}
 
To use GPT-3 to code texts, we provide it with a specific prompt designed to teach GPT-3 the coding process. This prompt varies from application to application, as the coding method depends on the specific concepts being coded. Throughout these applications, our goal is to give GPT-3 as little guidance as possible to demonstrate its flexibility and efficiency in learning how to act as a coder. In providing GPT-3 with these prompts, we discovered that the LM responded quite similarly across various versions of our guidance, and that it required only two or three coded examples to perform well on these tasks. For additional information on the process of engineering these prompts, see the Online Appendix.

After giving GPT-3 these prompts and observing how it codes a set of data, we compare that coding to a corresponding set of codes generated by humans. This allows us to directly compare the performance of GPT-3 to human coders. In the case of our last application, we also compare our results to a SML procedure. We make these comparisons based on coding agreement as well as efficiency (in terms of time and cost to code with other techniques).

\definecolor{myblue}{RGB}{76, 146, 195}
\definecolor{mygreen}{RGB}{86, 179, 86}
\definecolor{myred}{RGB}{222, 82, 23}
\definecolor{mypurple}{RGB}{169, 133, 202}
\definecolor{myorange}{RGB}{254, 153, 62}
\newcommand{\colorins}{\textcolor{myblue}}
\newcommand{\colorenu}{\textcolor{myorange}}
\newcommand{\colorexe}{\textcolor{mygreen}}
\newcommand{\colorcla}{\textcolor{myred}}

\newcommand{\figurefont}{7}

\begin{figure*}
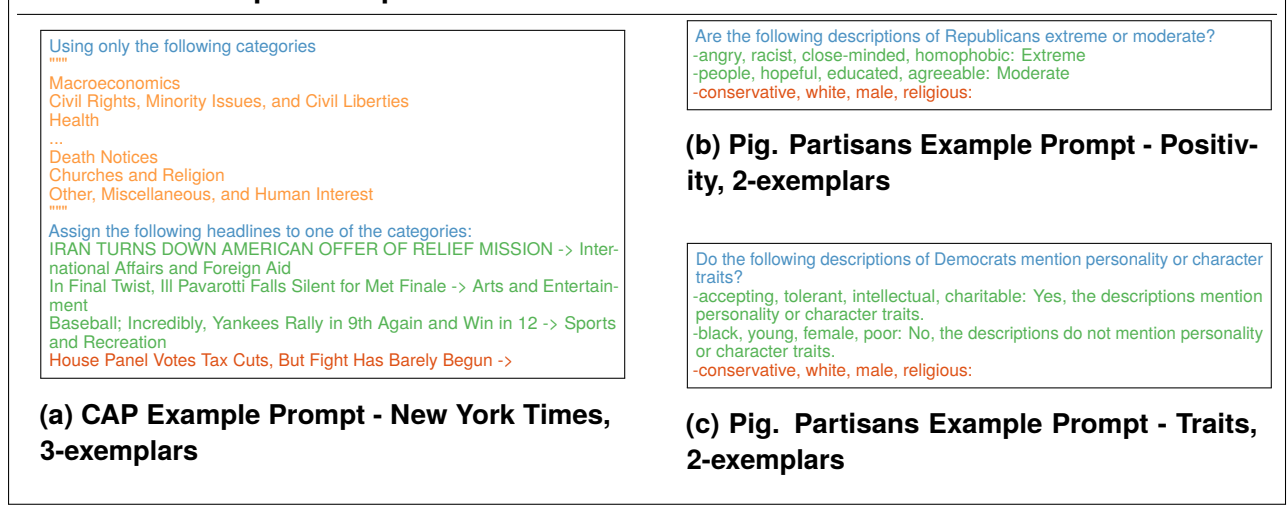

\centering
\begin{minipage}{.45\linewidth}
    \fbox{%
  \parbox{\textwidth}{
  \fontsize{\figurefont}{\figurefont}\selectfont
    \colorins{Using only the following categories}
    \colorenu{\\"""\\
    Macroeconomics\\
Civil Rights, Minority Issues, and Civil Liberties\\
Health\\
...\\
Death Notices\\
Churches and Religion\\
Other, Miscellaneous, and Human Interest\\
    """}
    \colorins{\\Assign the following headlines to one of the categories:}
    \colorexe{\\IRAN TURNS DOWN AMERICAN OFFER OF RELIEF MISSION -> International Affairs and Foreign Aid\\
     In Final Twist, Ill Pavarotti Falls Silent for Met Finale -> Arts and Entertainment\\
    Baseball; Incredibly, Yankees Rally in 9th Again and Win in 12 -> Sports and Recreation}
    \colorcla{\\House Panel Votes Tax Cuts, But Fight Has Barely Begun ->
    }
  }%
}
\vspace{0.2in}

    \subcaption{CAP Example Prompt - New York Times, 3-exemplars}
    \label{fig:capprompt}
\end{minipage}
\hspace{.05\linewidth}
\begin{minipage}{.45\linewidth}
    \fbox{%
  \parbox{\textwidth}{
  \fontsize{\figurefont}{\figurefont}\selectfont
    \colorins{Are the following descriptions of Republicans extreme or moderate?}
    \colorexe{\\-angry, racist, close-minded, homophobic: Extreme\\-people, hopeful, educated, agreeable: Moderate}
    \colorcla{\\-conservative, white, male, religious:}
  }%
}
\vspace{0.2in}

    \subcaption{Pig. Partisans Example Prompt - Positivity, 2-exemplars}
    \label{fig:promptpositivity}
\vspace{10pt}
    \fbox{%
  \parbox{\textwidth}{
  \fontsize{\figurefont}{\figurefont}\selectfont
    \colorins{Do the following descriptions of Democrats mention personality or character traits?}
    \colorexe{\\-accepting, tolerant, intellectual, charitable: Yes, the descriptions mention personality or character traits.\\
    -black, young, female, poor: No, the descriptions do not mention personality or character traits.}
    \colorcla{\\-conservative, white, male, religious:}
  }%
}
\vspace{0.2in}

    \subcaption{Pig. Partisans Example Prompt - Traits, 2-exemplars}
    \label{fig:traits}

\end{minipage}
    \caption{Example Prompts}
  \label{fig:exampleprompts}
\end{figure*}

We construct our prompts by providing \colorins{instructions}, \colorenu{categories} (if necessary), \colorexe{exemplars} (labeled examples of the task), and then the \colorcla{text to classify}. We then compute GPT-3's probabilities for the next token over its vocabulary and select the token with the highest probability as the model's coding choice. For color-coded examples of prompts, see Figure \ref{fig:exampleprompts}.

We evaluate GPT-3's coding performance using various intercoder agreement measures between GPT-3's codes and the codes generated by humans we hired to code the same texts. These are as follows: 

\subsection{Intraclass correlation (ICC)} 
Intraclass correlation measures inter-coder agreement among human coders using numerically ordered, (quasi-) continuous values in their coding (e.g., rating a text by some characteristic on a 1-5 scale). ICC scores are between -1 and 1 and are typically interpreted as follows: $<0.5$ = poor inter-coder agreement, $0.5-.75$ = moderate agreement, $0.75-0.9$ = good, and $>0.9$ = excellent \citep{Cicchetti1994,Koo2016}.

\subsection{Joint probability of agreement}
For tasks with un-ordered, categorical codes, we use two different measures. The first, joint-probability of agreement, measures the probability of any two coders agreeing. In the 2-coder case, where one of the coders is ground truth, this reduces to raw accuracy. Joint probability agreement ranges from 0 to 1. Between two coders, it is calculated as follows: $
\frac{1}{N} \sum_{i=1}^{N} \mathbbm{1}(y_{1,i} = y_{2,i})
$, where $N$ is the number of instances being coded, and $y_{1,i}, y_{2,i}$ are the first coder's and the second coder's respective codings of instance $i$. In the case of $K$ coders, the joint probability agreement is the mean of the pairwise agreements.

\subsection{Fleiss' kappa}
Fleiss' kappa measures the degree to which the proportion of agreement among coders exceeds the agreement of fully random coders \citep{Fleiss1971,Fleiss2003}. Used specifically to quantify intercoder agreement for categorical data, this measure ranges from $-1$ to $1$. When $\kappa = 0$, it means that the two raters agree at a rate not better than chance. $\kappa < 0$ means increasing agreement worse than chance, and $\kappa > 0$ means increasing agreement greater than chance.

\section{Experiments}

We consider GPT's capacity to serve as a coder using data from four datasets: Pigeonholing Partisans (PP), New York Times Headlines (NYT), Congressional Hearings (Congress), and The Guardian Populism (TGP). We chose these datasets to maximize differences in coding tasks as a means of exploring GPT-3's limits. These four applications vary in the difficulty of the coding task, the domain (or topic) of the coding, the structure of the texts, and measurement of the coded variable (ordinal, categorical, binary, etc.).

\subsection{Pigeonholing Partisans (PP)} \label{pigeonholingpartisans}

We first consider the ability of GPT-3 to act as a coder with data on Americans' stereotypes of Republicans and Democrats \citep{busby2019}. These data, collected in 2016, asked individuals to list four words or phrases that described typical supporters of the Democratic and Republican Parties.\footnote{More methodological details can be found in published discussions of this work. See \citep{busby2019}.} 
This procedure is common in psychological studies of stereotypes \citep{devine1989, eaglymladinic1989}, and allows survey takers to describe partisans in their own words 
This dataset is too small for other kinds of automated coding and an ideal way to consider how well GPT-3 can classify texts without extensive training sets.

To evaluate how well GPT-3 can serve as a coder on these kinds of short, open-ended texts, we recruited 2873 human coders through the survey platform \textit{Lucid} \citep{coppock2019} to code a total of 7675 descriptions of partisans. Each description was coded at least three times by a random set of coders, who were given minimal instructions for coding the texts.\footnote{These texts include those created by human respondents in the original data as well as texts created by GPT-3 and discussed in other, published work \citep{Arglyeetal}. That work indicates that human respondents cannot distinguish between the two kinds of statements.} As such, the coders in this study should be considered "lightly trained" rather than rigorously instructed on the coding.

Coders rated the texts along five dimensions: (1) positivity (general positive/negative valence), (2) extremity (extreme or moderate quality of the words), and whether the text mentioned (3) character or personality traits, (4) government or policy issues, or (5) social groups. Each of these domains is important to the theoretical ideas of the original work on partisan stereotypes \citep{busby2019, busby2021}. 

After the human coding process was complete, we asked GPT-3 to complete a series of coding tasks on all 7675 texts directly analogous those completed by humans. Next, we examined how closely GPT-3 follows individual human coders and human coding in the aggregate, along with how closely humans followed each other.

\begin{figure}
\includegraphics[trim=0 5 65 60, clip, width=.8\textwidth]
{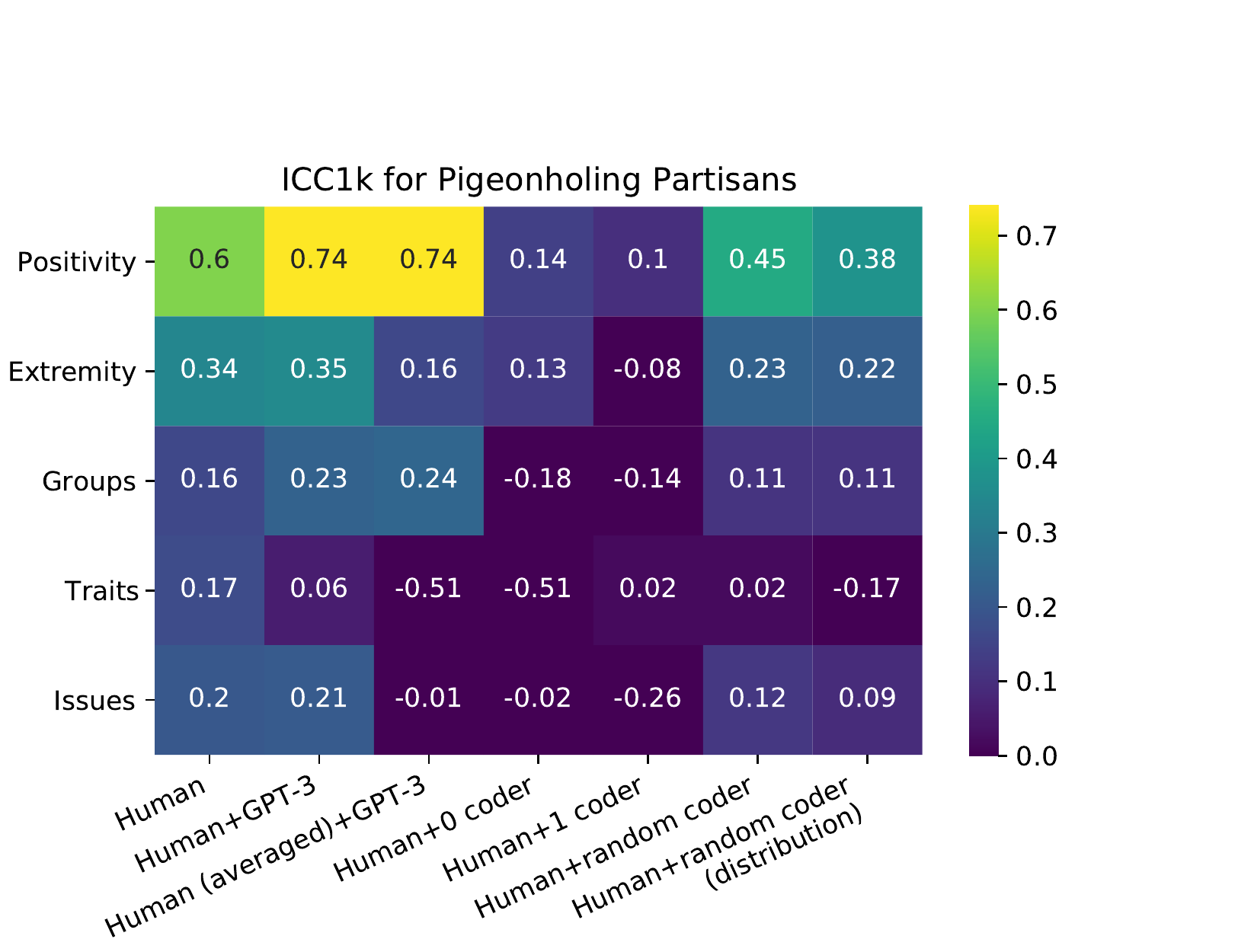}
\centering
\caption{PP ICC1k: Note that including GPT-3 in the class of considered coders increases ICC1k in coding for all attributes except ``Traits''. The opposite happens when including other, simulated coders.}
\label{fig:icc_lucid}
\end{figure}

To that end, we calculated ICC scores with these data (Fig.~\ref{fig:icc_lucid}). As coders are randomly assigned to texts and not all texts are scored by the same coders, we use ICC1k, which accounts for this structure \citep{Shrout1979}. 

Our focus here is on the increase or decrease in ICC when GPT-3's codes are added to the three human codes. If GPT-3 improves the reliability of the coding, ICC should improve. If it does not offer this benefit, the ICC score should stay the same or decrease. We also compare adding GPT-3's scores to adding simulated scores to ensure that the addition of another coder by itself does not drive what we observe: (1) a coder who codes all texts as 0 (lacking the attribute), (2) a coder who codes all texts as 1 (containing the attribute), (3) a coder who codes randomly, and (4) a coder who codes all texts randomly, but with the same overall distribution as GPT-3's predictions. We also consider the ICC values when comparing GPT-3's codes to the average of the human coders (rather than individual coders separately).

The statistics in Figure \ref{fig:icc_lucid} suggest that adding GPT-3 as a coder adds a great deal to reliability for two measures (positivity, groups), slightly increases reliability of the coding for two others, (extremity, issues), and reduces reliability in one (traits). Notably, this last area is where human coders correlated the least with each other (correlations between human coders on this domain ranging from 0.07 to 0.08) and may represent a fundamentally challenging task. 

There is also a stark difference between adding GPT-3 and adding each of the simulated coders. We conclude that the boost in ICC from GPT-3 is not due to simply adding another coder. Furthermore, since adding GPT-3's outputs to the human outputs generally either increases or maintains ICC across each attribute, we conclude that GPT-3 achieves human or better performance at this task. Importantly, achieving this level of performance required neither coding a large-scale dataset (on the order of tens of thousands or more) nor a large, labeled set of training data for the language model. 

\subsection{Comparative Agendas Project (CAP)} \label{cap}



For a different application of GPT-3 as a coder, we during to the Comparative Agendas Project (CAP) system of coding. CAP provides a coherent framework for documenting media and government attention to various policy issues in a comprehensive set of policy domains 
\citep{baumgartner2019comparative}. CAP datasets aim to be comprehensive, transparent, and replicable \citep{bevan2019gone}, with many housed at the CAP website (www.comparativeagendas.net). More than 200 scholars have used CAP to test a vast range of empirical political science theories across more than a dozen countries \citep{walgrave2019comparative}.

The CAP master codebook moves beyond the simple coding of the PP data, spanning at least 21 major categories (with others added for some specific applications). In order to succeed here, GPT-3 must produce a high probability for one of a large, unordered, pre-specified set of tokens that corresponds to the specific content of the input data.
    
Prior efforts to automate coding in the CAP framework have met limited success \citep{karan2016analysis, hillard2008computer, purpura2006automated, sevenans2014automated, sebHok2021multiclass}. Sebok and Kacsuk \citep{sebHok2021multiclass} are able to achieve an 80\%+ F1 score on average across categories, but this is reported after culling over 40\% of their dataset due to difficulty of classification. We, on the other hand, provide scores given full coverage of the dataset. Reported performance in various approaches is substantially lower than this (accuracies near or below 50\%) for dictionary methods, less efficient SMLs, corpora with less training data, or in specific hard-to code categories, which upper limit our average accuracy exceeds. Again, the highest performing outcomes are achieved by setting rejection thresholds (for ambiguous texts or cases where humans or models disagree) and either sacrificing coverage or targeting human coders to uncertain cases \citep{karan2016analysis, sebHok2021multiclass}. We achieve our without dropping cases, using multiple models, human disambiguation of difficult cases, and extensive labeled training data. 
    
To account for class imbalances and differences in baseline probabilities of different tokens, we normalize the probability distributions in a manner similar to \citep{zhao2021calibrate}. We estimate GPT-3's bias towards a category as the total weight given to each category over a balanced validation set, divide each category probability by GPT-3's bias towards it, and normalize to sum to 1. We found that this produced modest accuracy boosts of 4-5\%. If a small validation set is available, we recommend this calibration technique; however, results were qualitatively the same without this calibration.

We consider two data sources that have previously been coded using the CAP framework - coding of U.S. Congressional hearing summaries and the \textit{New York Times} front page. We conducted our coding with GPT-3 separately for each of these applications.
    
\begin{figure}
\includegraphics[scale = 0.8]{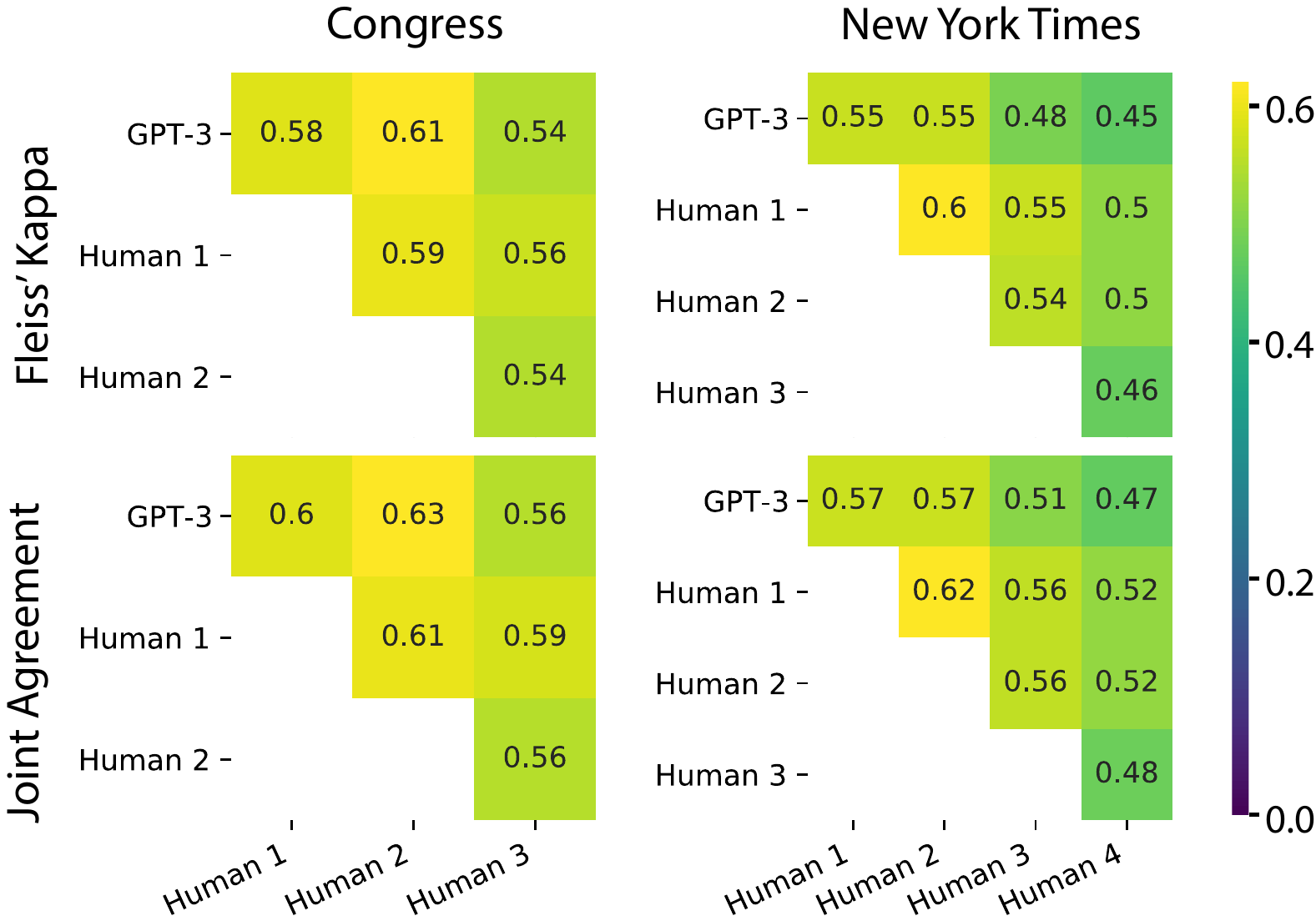}
\centering
\caption{Two measures of GPT-3's agreement with human coders compared with humans' agreement with human coders, across two datasets.}


\label{fig:jointfleiss}
\end{figure}
\subsubsection{CAP: Congressional Hearing Summaries (Congress)}
\label{subsubsec:congress}
        
The Congressional Hearing corpus contains the  \textit{Congressional Information Service} summary of each U.S. Congressional hearing from 1946 to 2010. These summaries were read by human coders and assigned to CAP classifications. We hired and trained three human coders for this application, providing them with the same instructions outlined in the CAP codebook. This allows us to compare how different human coders and GPT-3 compare to one another (which is not possible with the original data, given that it lacks scores from multiple coders). We gave GPT-3 the full summary text, making the coding task is highly comparable between the humans and GPT-3. All results are reported for $n=326$ texts, which constitutes 16 texts for each category minus 10 for incompleteness in the human codes. We used a random subset of the dataset of over 10,000 texts for this application. 




Figure \ref{fig:jointfleiss} presents our comparison of GPT-3's and the humans' codes. Both our intercoder agreement metrics tell the same story, and imply a finding that holds across metrics: GPT-3 correlates with each human just as well as or better than the humans correlate with each other. Note that the highest joint agreement (.63) and highest Fleiss' kappa (.61) both occur between GPT-3 and Human 2.

\begin{figure*}[t]
\includegraphics[trim=50 280 50 50, clip, width=17cm]
{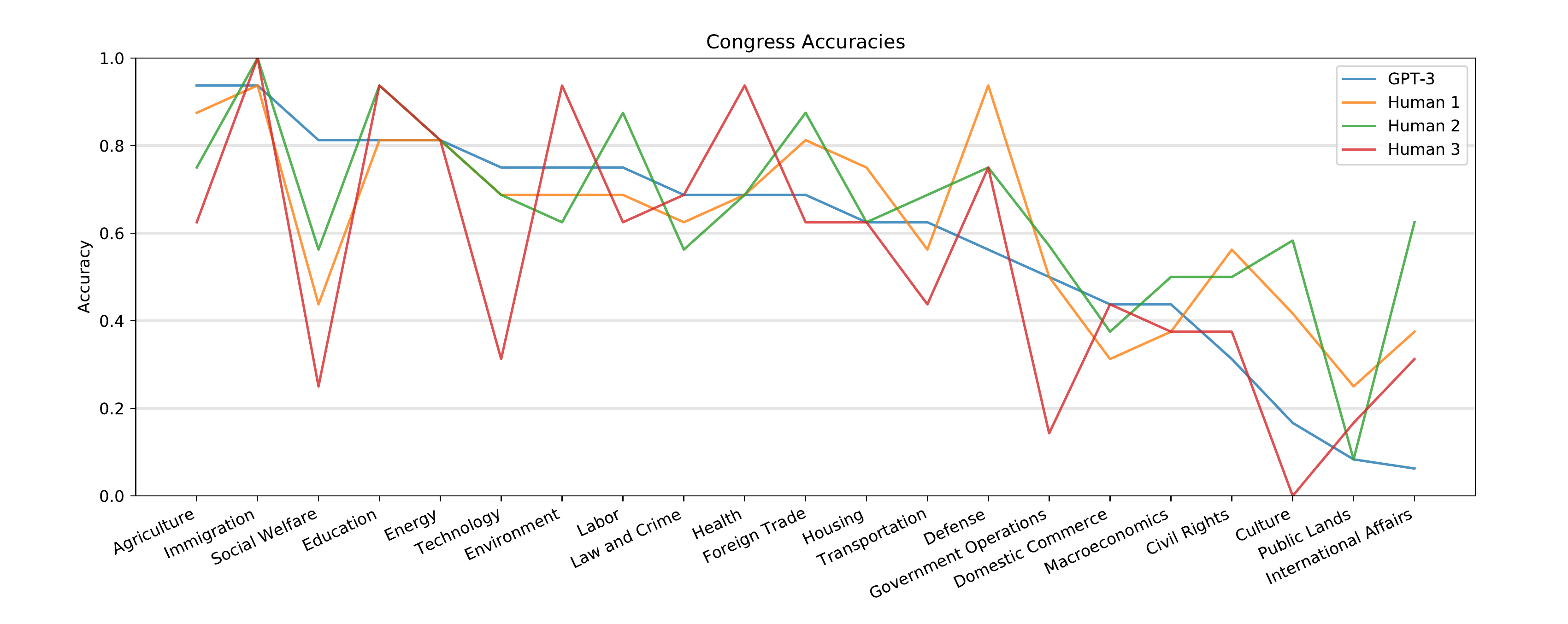}
\centering
\caption{Congress Accuracy by Coder: Treating the original dataset's code as ``ground truth'', and sorting categories in descending order based on GPT-3's score, note how noisy the performance of the human coders is. There is only 1 category that all humans score strictly better on (International Affairs).}
\label{fig:congressacc}
\end{figure*}

Despite there being no real ground truth for this task, we visualize ``accuracy'' statistics based on the original dataset's single coder as provided by CAP (Figure \ref{fig:congressacc}). The lack of ground truth is validated by a great deal of human disagreement, as the figure makes clear. We see the accuracy for each coder, with categories sorted in order of GPT-3's accuracy. Interestingly enough, GPT-3 seems to do better at categories that humans do better at, and worse at the categories that humans fail at. Overall, the accuracies were 60\% for GPT-3, compared to 63\%, 66\%, and 55\% for the three human coders respectively.

The high joint agreement and Fleiss' kappa between GPT-3 and the human coders, as well as the similar accuracies across categories, demonstrate GPT-3 performance on-par with humans on this dataset. Given the efficiency gains from using GPT-3, such as lower costs in training coders and scalability to a large number of texts, we suggest that this gives additional evidence in favor of the usefulness of LMs as coders.







\subsubsection{CAP: \textit{New York Times} Front Page Dataset (NYT)}
\label{subsubsec:NYT}

The second CAP dataset we use is the \textit{New York Times} Front Page Dataset, generated and contributed by Amber Boydstun \citep{boydstun2013making}. The dataset includes 31034 front page \textit{New York Times} headlines from 1996 - 2006, along with the policy category label assigned by trained human coders. The categories are adapted for media use, and so include 28 primary classification categories. For this application, we randomly sampled 20 texts from each of the 28 categories to be coded by four human coders and GPT-3. All results are reported for the correspondent set of $n=560$ texts.

The original human coders were instructed to read the headline \textit{and the first three paragraphs of the article.} In our work, GPT-3 is only provided the headline, because the full article text is not available in the public data. To control for this difference in available information, we also hired four human coders complete an identical classification task to GPT-3, considering only the article headlines.



\begin{figure*}
\includegraphics[trim=50 230 50 50, clip, width=17cm]
{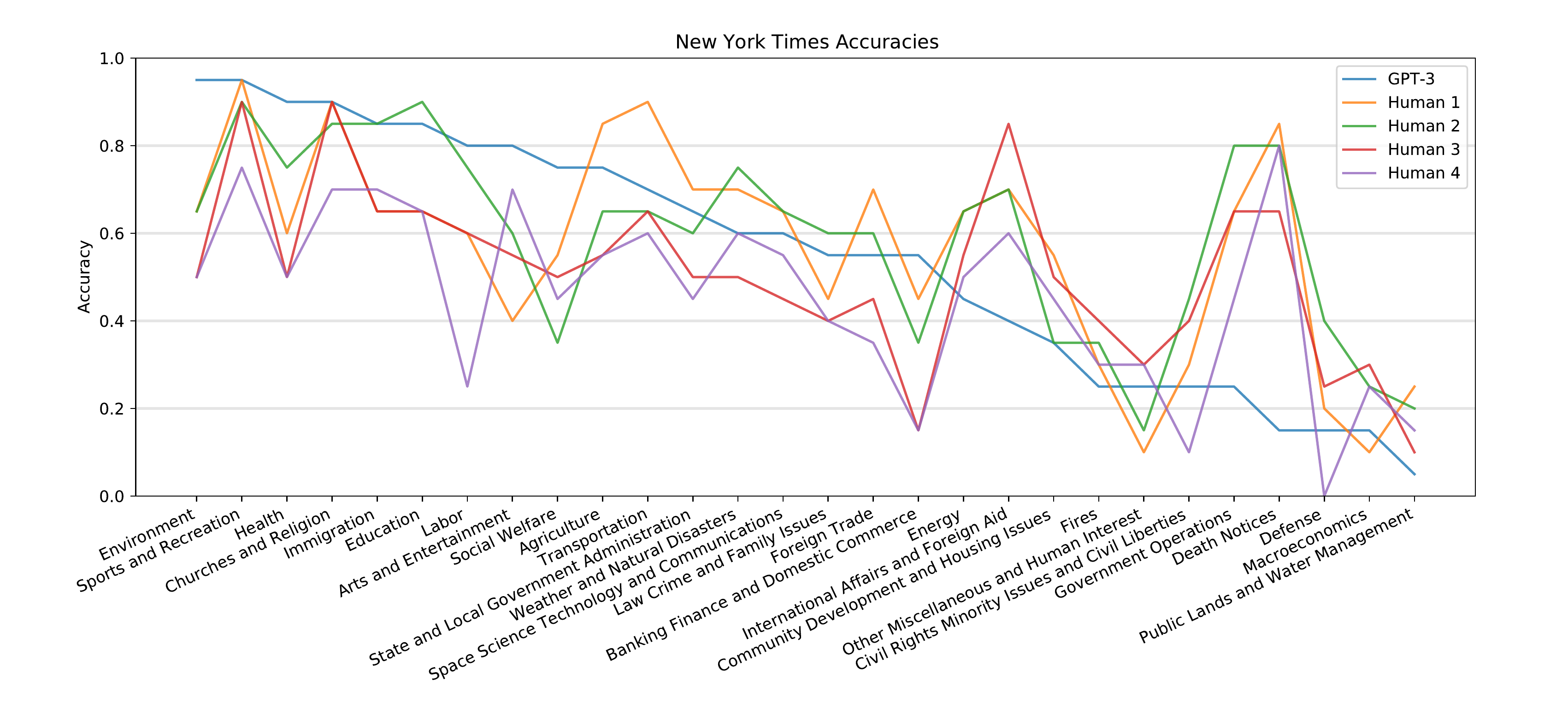}
\centering
\caption{New York Times Accuracy by Coder: Treating the original dataset code as ``ground truth'', and sorting categories in descending order according to GPT-3's score, note how noisy the humans' coding is. Clearly some areas are easier for human coders (e.g., Death Notices) and some are easier for GPT-3 (e.g., Environment).}
\label{fig:nytacc}
\end{figure*}

Since the structure of the NYT data is the same as the Congress data, we use the same kind of analyses. For both joint agreement and Fleiss' kappa (Figure \ref{fig:jointfleiss}), GPT-3 agrees with the humans about as much as they agree with each other. GPT-3's total accuracy was 55\%, compared to 57\%, 59\%, 51\%, and 45\% for the four humans respectively. We also notice a strong trend between GPT-3's accuracy and the humans accuracy per category (Figure \ref{fig:nytacc}). Unlike Congress, however, there are 3 categories for which the humans all perform better than GPT-3: ``International Affairs and Foreign Aid,'' ``Government Operations,'' and ``Death Notices.'' On the other hand, GPT-3 performs better than humans at some other categories: ``Environment,'' ``Health,'' and ``Labor.'' Overall, these results again demonstrate that GPT-3 generally achieves on-par performance with humans.

\subsection{\textit{The Guardian} Populism (TGP)} \label{tgp}

For our final application, we consider how GPT-3 codes a multifaceted concept - populism. While disagreement exists about the meaning of this term, many scholars have gravitated towards a definition that populism is a discourse that describes politics as a struggle between the virtuous will of the common people and some evil, conspiring elite \citep{mudde2017, Hawkins2009}.\footnote{This approach is sometimes called the "ideational" approach to populism} Coding for populism requires a process of marking the presence of a reference to the common people \textit{and} an evil elite. As such, existing studies have primarily relied on extensively trained human coders that are instructed on how to holistically code an entire text, examining it for references of both of these components (for an example of such a coding process, see \citep{busbygublerhawkins2019}).

Here we draw on a large dataset of short statements coded for populism. In the Fall of 2018, \textit{The Guardian} created a series of articles on populism. At the end of one article, readers were invited to participate in a related survey on populism - over 20,000 individuals from more than 100 countries completed this survey. One question on this study asked respondents to discuss who or what was responsible for a pressing political problem in their country; two intensively trained human coders evaluated 4,000 of these texts and indicated if they did or did not contain populism. The process of training these coders involved initial instruction on a set of unrelated texts, repeated sessions to correct mistakes and clarify the coding process, and a review of the human codes \citep{busbycarlinhawkinslittvay}. Unlike the preceding studies, then, this application involves comparisons to highly trained human coders. 

These data also allow for a comparison to SML methods, as about 16,000 texts were not coded by the human coders. As discussed below, we employ a SVC method to code the full set of texts and compare the performance of this technique to coding by GPT-3. We therefore compare the coding produced by GPT-3 on the set of human coded texts and in comparison to the SML approach. In each case, the coders (human or otherwise) generated a code of 1 when the text contained a populist statement and 0 when it did not. To be regarded as populist, the text needed to contain both a reference to the virtuous or good people and some kind of malicious elite group. \footnote{For more details on the human coding process, see other work explaining the codebook in more detail, such as \citep{busbycarlinhawkinslittvay} and \citep{busbygublerhawkins2019}}

We begin by comparing GPT-3's coding to the two human coders. As before, we calculated ICC scores to measure agreement between the coders. In contrast to the Pigeonholing Partisans data, the same two coders and GPT-3 coded all of the texts. We therefore use ICC3k which is designed for these kinds of comparisons \citep{Shrout1979}. For these comparisons, We had GPT-3 code a random sample of 1,300 of the 4,000 texts coded by humans.

[ADD FIGURE HERE]

Figure [FILL IN] shows the ICC statistics with GPT-3, the human coders, and the same types of simulated coders show in Section \ref{pigeonholingpartisans}. With these calculations, we find that GPT-3 performs well, although not quite as well as a thoroughly trained coder. The ICC statistic for the two human coders was 0.81, indicating high levels of agreement. Adding GPT-3 as a coder reduces this somewhat to 0.77, but this still indicates good agreement between the human coders and GPT-3. In contrast, adding one of the simulated coders dramatically reduces the ICC statistics. We take this as evidence that GPT-3 creates codes that are generally comparable to highly-trained human coders, with far less expense and training.

To compare GPT-3's performance to a supervised baseline, we fit a bag-of-words SML model on the populism data, using 3000 instances for training and 1000 instances for validation at a time. With this approach, the SML coding matched the human populism codes with an accuracy of 86 percent. Meanwhile, with only 4 coded examples, GPT-3 matched the human populism codes 79 percent of the time. While the SML baseline outperforms GPT-3 by about 7 percentage points, it does so at the cost of 3000 labeled examples. Given the drastically lower costs of coding with GPT-3 - in the case, the requirement of hiring, training, and supervising coders to classify 4,000 texts - we again see this as evidence of the value of GPT-3 as a coding tool for the social sciences.

\section{Ethics and Bias}
Our results suggest that GPT-3 can automate specific coding tasks comparably to human coders and SML coding methods. However, much work remains to bring this possibility to full fruition. For example, LMs reflect and even amplify pathological human biases contained in their training data \citep{Zhao2017}, raising concerns about their use for coding. Much work has aimed to quantify and reduce this bias \citep{Bordia2019, Qian2019}. However, while LMs exhibit bias, it is a known, invariant, and quantifiable property, whereas individual humans' biases are typically unknowable and far more difficult to quantify. We submit that the ability to recognize and actively compensate for the coder's probable biases is more important than the magnitude of the biases themselves. Conversely, if a LM can be conditioned or fine-tuned into holding specific biases rather than others, then it could emulate specific heterogeneous groups of coders for a richer, more diverse, and representative coding than what we present in this paper. 

In that sense, we suggest that bias in coders is an omnipresent problem in coding for the social sciences. Here again, LMs provide a way to evaluate and account for those problems. We encourage other researchers interested in and employing LMs in their coding to use this tool to improve the accuracy and inclusivity of their coding and not simply their efficiency.

\section{Conclusion}
With four dramatically different sources of data, we have demonstrated that LMs can be used to code social science datasets more efficiently and as accurately as existing human or SML techniques. Fine-grained analysis shows that GPT-3 can match the performance of human coders on average across small and large datasets; with both ordinal and categorical codes; and on tasks of varying complexity. In some cases, it even outperforms humans in increasing intercoder agreement scores, often with no more than 3 exemplars. 
 
We suggest that these results indicate the promise of LMs (and other tools like them) for research in the social sciences. Our analyses are a first step in this direction, but tools like GPT-3 offer low-cost ways to process and evaluate large text corpora from various sources. They also allow researchers to perform these tasks while still using their theoretical and substantive knowledge of the topic at hand to guide the machine learning tools. As such, we view this as a productive synergy of human and computer components to generate an outcome more accurate and efficient than either element on its own. Given the turn of the social sciences towards promising new terrains of text and other complex data, LMs and other related tools offer great promise for nearly every domain of the social sciences.


\bibliography{references}

\newpage

\appendix
\section{APPENDIX} 
\subsection{Prompt Engineering} \label{app:promptengineering}

One important part of this project is the way that we provide information or the context to GPT-3 to help it learn the coding process. As noted in the text, our overarching goal was to make this instruction as minimal and flexible as possible to evaluate GPT-3's potential without dramatic changes to the LM. In doing so, we learning a number of important lessons about giving GPT-3 information about the coding scheme and process. In this section, we seek to explain our prompt engineering protocol so that our results can be replicable and generalizable to other datasets and domains. We include both decisions made without conducting any experiments and those made by conducting experiments.

Some elements of prompt engineering seem to matter a great deal, and some seem to matter not much. Of all the sections of this paper, we spent the most time on this one, and ran the most experiments to fill it. Despite this, and the fact that slight changes to individual prompts on formatting cause significant changes to the probability distribution over tokens, we found that in the aggregate, prompt engineering tends to not make much of a difference in GPT-3's performance as a coder.

\begin{figure}[b]
\includegraphics[scale=0.5]{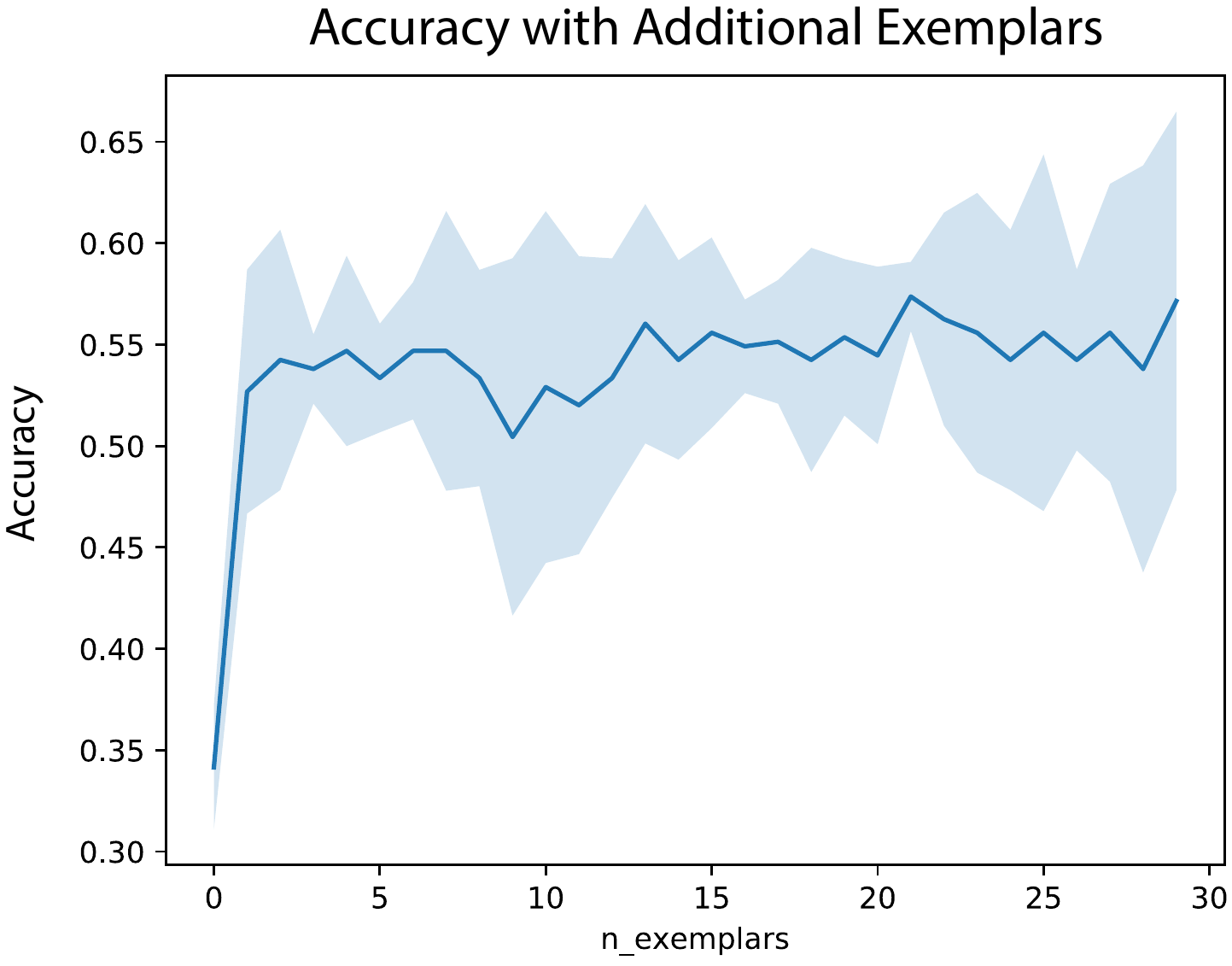}
\caption{Increasing number of exemplars up to 30 shows no improvement past 2 or 3. This experiment was done on the NYT dataset.}
\label{fig:n_exemplars_accuracy}
\end{figure}

In this process, one has to be mindful of where the prompt ends and what next token is being modeled. Since generative language models sample one token at a time, we needed to be able to sample a unique first token (usually, a unique first word) for each category we attempt to model. For example, ``very positive'' and ``very negative'' both start with the token ``very,'' so it would be impossible for us to compare the two categories with a single token sample. Fortunately, all of our categories started with unique first tokens, but will not be true for all future applications of LMs as coders.

Another choice impacting results was the presentation of categories in the question format of the PP data. Specifically, GPT-3 performed significantly worse when asked to respond to a question with the tokens ``yes'' or ``no'' than when the choice was between substantive alternatives, such as ``extreme'' vs ``moderate'' or ``positive'' vs. ``negative''. For the other three attributes, we found that restating the objective after the ``yes'' or ``no'' (e.g., ``Yes, mentions personality or character traits'') substantially helped. These were the only prompt variations attempted for the PP dataset.

Other elements seemed to have minimal impact, like the number and type of exemplars. While we know that more labeled training data significantly improves SML performance \citep{collingwood2012tradeoffs}, it was unclear ahead of time whether more labeled exemplars to GPT-3 will achieve the same.  In theory, more exemplars could more firmly teach the model the format of the task, and every marginal quality exemplar could help the model refine its understanding of the distributions of categories that the examples belong to. 

As shown in Figure A.\ref{fig:n_exemplars_accuracy}, we find that one exemplar performs much better than none, but there is little gain in accuracy achieved by providing more than 2 or 3 exemplars. We also conducted extensive experiments testing different classes of exemplars (more or less difficult to classify, in the spirit of active learning); this also seemed not to matter (See Appendix \ref{app:exemplars} for more details).

We also tried many variations on the prompt format, including: surrounding categories in quotes; using slashes, pipes, and other delimiters to separate exemplar headlines from their respective categories; providing lists of example headlines for each category in parentheses right next to the category; new lines in specific places making boundaries between exemplars clearer; and other general rephrasing. None of these changes resulted in a marginal accuracy less than 50\% or greater than 57\%. This demonstrates a relative stability of the information retrieval process, allaying some concerns that minor changes in wording or punctuation will radically alter coding accuracy.

For all of our final prompts used, please refer the following section.

\subsection{Prompts For Each Task}
\label{app:prompts}

\subsubsection{Pigeonholing Partisans}

\begin{itemize}

\item 
\textbf{Positivity}: 
\begin{quote}
Are the following descriptions of PARTY positive or negative?\\-agreeable, reasonable, understanding, cooperative: Positive\\-angry, bigoted, racist, homophobic: Negative
\end{quote}

\item 
\textbf{Groups}: 
\begin{quote}
Do the following descriptions of PARTY mention social groups?\\-Christian, privileged, young, white: Yes, mentions social groups.\\-apathetic, agreeable, pro-environment, political: No, doesn't mention social groups.
\end{quote}

\item 
\textbf{Traits}: 
\begin{quote}
Do the following descriptions of PARTY mention personality or character traits?\\-accepting, tolerant, intellectual, charitable: Yes, mentions personality or character traits.\\-black, young, female, poor: No, doesn't mention personality or character traits.
\end{quote}

\item 
\textbf{Extremity}: 
\begin{quote}
Are the following descriptions of PARTY extreme or moderate?\\-angry, racist, close-minded, homophobic: Extreme\\-people, hopeful, educated, agreeable: Moderate
\end{quote}

\item 
\textbf{Issues}: 
\begin{quote}
Do the following descriptions of PARTY include government or policy issues?\\-aging, religious, accepting, patriotic: No, doesn't include government or policy issues.\\-abortion, medical marijuana, gun control, anti-sexism: Yes, includes government or policy issues.
\end{quote}

\end{itemize}

\subsubsection{CAP}

\begin{itemize}
    \item 
\textbf{Congressional Hearings}:
\begin{quote}
Using only the following categories\\"""\\Macroeconomics\\Civil Rights\\Health\\Agriculture\\Labor\\Education\\Environment\\Energy\\Immigration\\Transportation\\Law and Crime\\Social Welfare\\Housing\\Domestic Commerce\\Defense\\Technology\\Foreign Trade\\International Affairs\\Government Operations\\Public Lands\\Culture\\"""\\Assign the following congressional hearing summaries to one of the categories:\\Extend defense production act provisions through1970. -> Defense\\FY90-91 authorization of rural housing programs. -> Housing\\Railroad deregulation. -> Transportation\\To consider Federal Reserve Board regulations and monetary policies after February 2016 report on monetary policy. ->'
\end{quote}

\item
\textbf{New York Times Headlines}
\begin{quote}
Using only the following categories\\"""\\Macroeconomics\\Civil Rights, Minority Issues, and Civil Liberties\\Health\\Agriculture\\Labor\\Education\\Environment\\Energy\\Immigration\\Transportation\\Law, Crime, and Family Issues\\Social Welfare\\Community Development and Housing Issues\\Banking, Finance, and Domestic Commerce\\Defense\\Space, Science, Technology and Communications\\Foreign Trade\\International Affairs and Foreign Aid\\Government Operations\\Public Lands and Water Management\\State and Local Government Administration\\Weather and Natural Disasters\\Fires\\Arts and Entertainment\\Sports and Recreation\\Death Notices\\Churches and Religion\\Other, Miscellaneous, and Human Interest\\"""\\Assign the following headlines to one of the categories:\\IRAN TURNS DOWN AMERICAN OFFER OF RELIEF MISSION -> International Affairs and Foreign Aid\\ In Final Twist, Ill Pavarotti Falls Silent for Met Finale -> Arts and Entertainment\\In Times Sq., a Dry Run for New Year\'s 2000 -> Arts and Entertainment\\House Panel Votes Tax Cuts, But Fight Has Barely Begun ->'
\end{quote}
\end{itemize}

\newpage
\subsection{Exemplar Types Experiments}
\label{app:exemplars}

We also explored whether some exemplars were better or worse at ``teaching'' the categories to the model. We considered that for a given category, an instance could be a better or worse exemplar. We might define this by a quantity we'll call its \textit{margin}: the difference between (1) the probability the model assigns to the correct category and (2) the highest probability of the probabilities for all the wrong categories. Thus, ``prototypical" exemplars would have high positive margin (model guesses right), ``ambiguous" exemplars would have margins with very low absolute values (model torn between multiple categories), and ``tricky" exemplars would have margins with very high negative values (model guesses wrong).   
In theory, prototypical exemplars could teach the model about the proper distribution of texts belonging to a category, ambiguous exemplars could teach the model about the boundaries between the distributions of each category, and tricky exemplars could correct the model's prior on categories by flagging common mistakes made in coding texts from that category's distribution.

\begin{figure}[h]
\includegraphics[scale = 0.5]{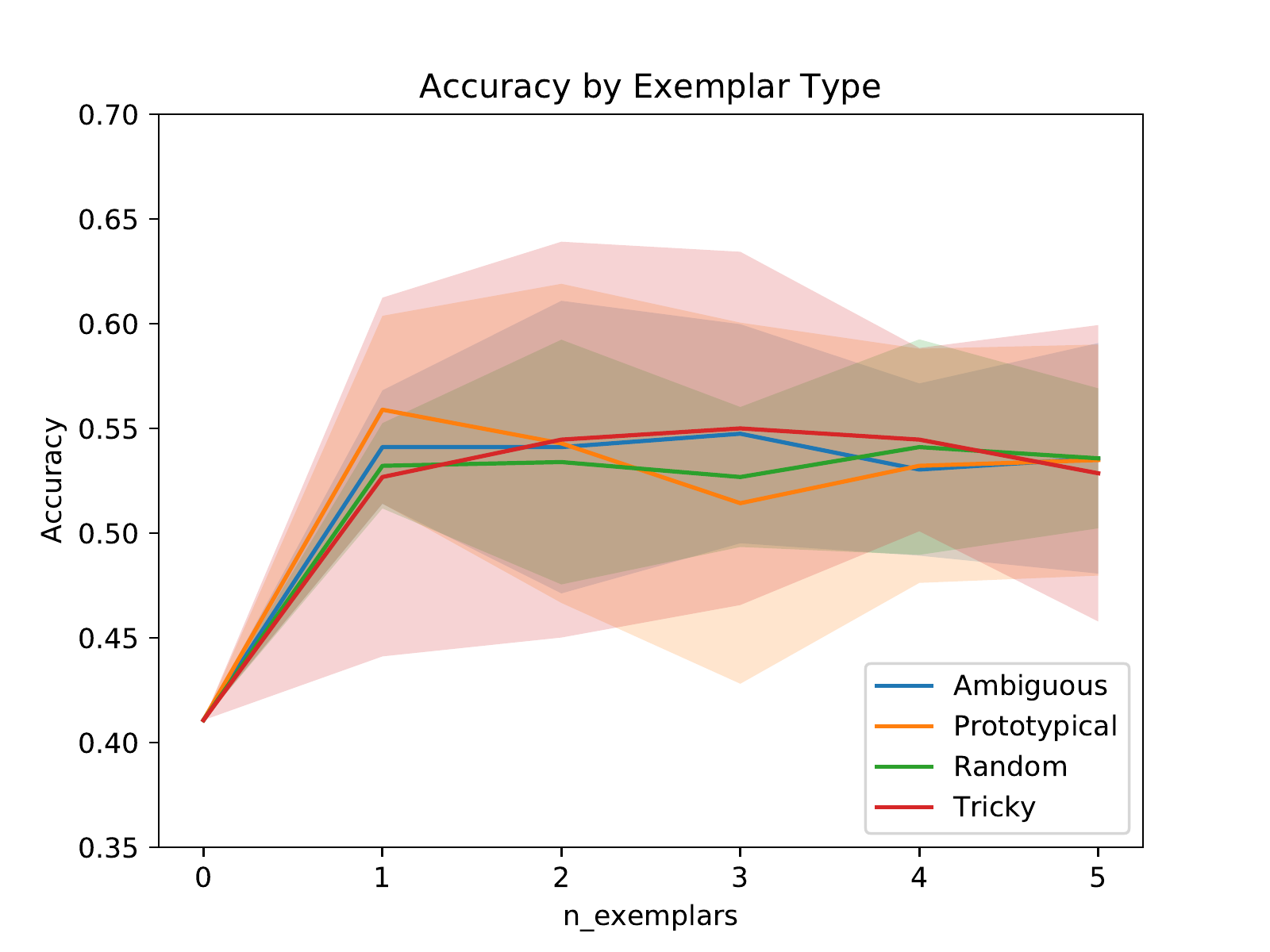}
\caption{Each class of exemplar considered does an equal amount to help the model's accuracy. This is surprising, and suggests that the model might learn nothing from the exemplars besides the format of the task.}
\label{fig:exemplar_types_accuracy}
\end{figure}

To answer this question empirically, we first randomly sample 90 candidate exemplars from each category. We then code each with the model given a set of 4 exemplars sampled randomly once and then held constant specifically for this task. Then we sort them by their margin and construct one set of each: prototypical, ambiguous, and tricky exemplars. Finally, we perform 5 trials where we classify 4 instances from each category using an increasing number of these sets of exemplars and measure performance. The results, in Figure A.\ref{fig:exemplar_types_accuracy}, demonstrate no discernible signal as to which kind of exemplar is best to present to the model in the context window. This is one bit of evidence that this dimension, of the prototypicality vs. ambiguity vs. trickiness of exemplars, is not at all determinative of a model's performance on a coding task, a dimension which is very important for active learning.


\end{document}